\documentclass[11pt,a4paper]{article}
\usepackage[]{main}
\usepackage{times}
\usepackage{latexsym}

\usepackage{bm}
\usepackage{amssymb}
\usepackage{graphicx}
\usepackage{subcaption}
\usepackage{soul}

\usepackage[normalem]{ulem}  %

\newif\ifcomments
\commentsfalse

\ifcomments
    \newcommand{\ian}[1]{\textcolor{olive}{\textbf{Ian: #1}}}
    \definecolor{iansnotecolor}{RGB}{255, 218, 181}
    \newcommand{\iansidenote}[1]{
        \todo[color=iansnotecolor, size=\footnotesize]{%
        [\textbf{Ian:}] #1}%
    }
    
    \newcommand{\amil}[1]{\textcolor{red}{\textbf{Amil: #1}}}
    
    \newcommand{\elahe}[1]{\textcolor{violet}{\textbf{Elahe: #1}}}
    \newcommand{\ellie}[1]{\textcolor{blue}{\textbf{Ellie: #1}}}

    \newcommand{\optional}[1]{\textcolor{purple}{\it #1}}
    \newcommand{\inlinetodo}[1]{\textcolor{red}{\textbf{TODO: #1}}}
    \newcommand{\done}[1]{\textcolor{gray}{\sout{#1} (done)}}
    
    \newcommand{\maybedelete}[1]{\textcolor{red}{\sout{#1}}}

\else
    \newcommand{\ian}[1]{}
    \newcommand{\amil}[1]{}
    \newcommand{\elahe}[1]{}
    \newcommand{\iansidenote}[1]{}
    \newcommand{\ellie}[1]{}
    
    \newcommand{\optional}[1]{}
    \newcommand{\inlinetodo}[1]{}
    \newcommand{\done}[1]{}
    
    \newcommand{\maybedelete}[1]{}

\fi

\aclfinalcopy %

\title{What Happens To BERT Embeddings During Fine-tuning?}

\author{Amil Merchant$^1$
\thanks{\quad Work done as member of the Google AI Residency program \url{https://ai.google/research/join-us/ai-residency/}} \quad Elahe Rahimtoroghi$^1$ \quad Ellie Pavlick $^{1,2}$ \quad Ian Tenney$^1$ \\ $^1$ Google Research \quad $^2$ Brown University \\ \texttt{\{amilmerchant, elahe, epavlick, iftenney\}@google.com}}
  
\date{}

\begin{document}
\maketitle

\begin{abstract}
While there has been much recent work studying how linguistic information is encoded in pre-trained sentence representations, comparatively little is understood about how these models change when adapted to solve downstream tasks. Using a suite of analysis techniques (probing classifiers, Representational Similarity Analysis, and model ablations), we investigate how fine-tuning affects the representations of the BERT model. We find that while fine-tuning necessarily makes significant changes, it does not lead to catastrophic forgetting of linguistic phenomena. We instead find that fine-tuning primarily affects the top layers of BERT, but with noteworthy variation across tasks. In particular, dependency parsing reconfigures most of the model, whereas SQuAD and MNLI appear to involve much shallower processing. Finally, we also find that fine-tuning has a weaker effect on representations of out-of-domain sentences, suggesting room for improvement in model generalization.

\end{abstract}

\ifcomments
\else
\fi

\section{Introduction}
The introduction of unsupervised pre-training for Transformer architectures \cite{Vaswani_Attention} has led to significant advances in performance on a range of NLP tasks. Most notably, the popular BERT model \cite{Devlin_BERT} topped the GLUE \cite{Wang_GLUE} leaderboard when it was released, and similar models have continued to improve scores over the past year \cite{Anonymous_ALBERT, Raffel_Exploring}.

Much recent work
has attempted to better understand these models and explain what makes them so powerful. Particularly, behavioral studies \citep[][\textit{inter alia}]{Goldberg_Assessing}, diagnostic probing classifiers \citep[][\textit{inter alia}]{Liu_Linguistic}, and unsupervised techniques \citep[][\textit{inter alia}]{Voita_The_Bottom} have shed light on the representations from the pre-trained models and have shown that they encode a wide variety of linguistic phenomena \cite{Tenney_What}.

However, in the standard recipe for models such as BERT \cite{Devlin_BERT}, after a Transformer is initialized with pre-trained weights, it is then trained for a few epochs on a supervised dataset. Considerably less is understood about what happens 
during this fine-tuning stage. 
Current understanding is mostly derived from observations about model behavior: fine-tuned Transformers achieve state-of-the-art performance but also can end up learning shallow shortcuts, heuristics, and biases \cite{Mccoy_Right, Mccoy_Berts, Gururangan_Annotation, Poliak_Hypothesis}. Thus, in this work, we seek to understand how the internals of the model--the representation space--change when fine-tuned for downstream tasks. We focus on three widely-used NLP tasks: dependency parsing, natural language inference (MNLI), and reading comprehension (SQuAD), and ask:
\begin{itemize}
    \itemsep0em
    \item What happens to the encoding of linguistic features such as syntactic and semantic roles? Are these preserved, reinforced, or forgotten as the encoder learns a new task? (Section~\ref{sec:probing})
    \item Where in the model are changes made? Are parameter updates concentrated in a small number of layers or are there changes throughout? (Section~\ref{sec:rep_change}) 
    \item Do these changes generalize or does the newfound behavior only apply to the specific task domain? (Section~\ref{sec:generalization}) 
\end{itemize}
We approach these questions with three distinct analysis techniques. Supervised probing classifiers \citep{Tenney_What, Hewitt_A_Structural} give a positive test for specific linguistic phenomena, while Representational Similarity Analysis \citep[RSA; ][]{Kriegeskorte_Representational} gives a task-agnostic measurement of the change in model activations. Finally, we corroborate the probing and RSA results with two types of model ablations--truncation and partial freezing--and measure their effect on end-task performance. 

Taken together, we draw the following conclusions. First, linguistic features are not lost during fine-tuning. Second, fine-tuning tends to affect only the top few layers of BERT, albeit with significant variation across tasks: SQuAD and MNLI have a relatively shallow effect, while dependency parsing involves deeper changes to the encoder. We confirm this by partial-freezing experiments which test how many layers \textit{need} to change to do well on each task and relate this to an estimate of task \textit{difficulty} with layer ablations. Finally, we observe that fine-tuning induces large changes on in-domain examples, but the representations of out-of-domain sentences resemble those of the pre-trained encoder.

\section{Related Work}

\paragraph{Base model} Many recent papers have focused on understanding sentence encoders such as ELMo \cite{Peters_Deep} and BERT \cite{Devlin_BERT}, focusing primarily on the ``innate" abilities of the pre-trained (``Base") models. For language models that report perplexity scores, behavioral analyses \cite{Goldberg_Assessing, Marvin_Targeted, Gulordava_Colorless,Ettinger_BertNot} have shown that they capture phenomena like number agreement and anaphora. For Transformer-based models \cite{Vaswani_Attention}, analyses of attention weights have shown interpretable patterns in their structure \cite{Coenen_Visualizing, Vig_Analyzing, Voita_Analyzing, Hoover_Exbert} and found strong correlations to syntax \cite{Clark_What}. However, other studies have also cast doubt on what conclusions can be drawn from attention patterns \cite{Jain_Attention, Serrano_Is, Brunner_On}.

More generally, supervised probing models (also known as diagnostic classifiers) make few assumptions beyond the existence of model activations and can test for the presence of a wide variety of phenomena. \citet{Adi_Fine, Blevins_Deep} tested recurrent networks for sentence length, word context, and syntactic properties, while \citet{Conneau_What, Jawahar_What} use another 10 probing tasks, including sensitivity to bigram shift. \citet{Tenney_What, Liu_Linguistic, Peters_Dissecting} introduced task suites that probe for high-level linguistic phenomena such as part-of-speech, entity types, and coreference, while \citet{Tenney_BERT} showed that these phenomena are represented in a hierarchical order in the different layers of BERT. \citet{Hewitt_A_Structural} also used a geometrically-motivated probing model to explore syntactic structures, and \citet{Wallace_Do} explored the ability of ELMo and BERT to faithfully encode numerical information.\footnote{See \citet{Belinkov_Analysis} and \cite{Rogers_Primer} for a survey of probing methods.}

While probing models depend on labelled data, parallel work has studied the same encoders using unsupervised techniques. \citet{Voita_The_Bottom} used a form of canonical correlation analysis \citep[PW-CCA; ][]{Morcos_Insights} to study the layer-wise evolution of representations, while \citet{Saphra_Understanding} explored how these representations evolve during training. \citet{Abnar_Blackbox} used Representational Similarity Analysis \citep[RSA; ][]{Laakso_Content,Kriegeskorte_Representational} to study the effect of context on encoder representations, while \citet{Chrupala_Correlating} correlated them with syntax. \citet{Abdou_Higher, Gauthier_Linking} also compared these representations to fMRI and eye-tracking data.

\paragraph{Fine-tuning} In contrast to the pre-trained models, there have been comparatively few studies on understanding the fine-tuning process. Initial studies of fine-tuned encoders have shown state-of-the-art performance on benchmark suites such as GLUE \cite{Wang_GLUE} and surprising sample efficiency. However, behavioral studies with challenge sets \cite{Mccoy_Right, Poliak_Hypothesis, Ettinger_Assessing, Kim_Teaching} have shown limited ability to generalize to out-of-domain data and across syntactic perturbations.

Looking more directly at representations, \citet{Aken_How} focused on question-answering models with task-specific probing models and clustering analysis. They found evidence of different stages of processing in a fine-tuned BERT model but showed only limited comparisons with the pre-trained encoder. \citet{Hao_Visualizing} explored fine-tuning from an optimization perspective, finding that pre-training leads to more efficient and stable optimization than random initialization. \citet{Peters_Tune} also analyzed the effects of fine-tuning with respect to the performance of diagnostic classifiers at various layers. \citet{Gauthier_Linking} is closest to our work: while focused on correlating representation to fMRI data, they also studied fine-tuning using RSA and the structural probe of \citet{Hewitt_A_Structural}, finding a significant divergence between the final representations of models fine-tuned on different tasks. By comparison, we seek a more general analysis of the internal representations of fine-tuned BERT models and also focus on how they change compared to the pre-trained Base model.

\section{Experimental Setup}

\paragraph{BERT}
In this paper, we focus on the effects of fine-tuning for the popular BERT architecture \cite{Devlin_BERT}. During pre-training, the model is initialized by training on masked language-modeling and next sentence prediction, with an unsupervised corpus from BooksCorpus \cite{Zhu_Aligning} and English Wikipedia. This model uses WordPiece tokenization \cite{Wu_Googles} and prepends input sentences with a \verb|[CLS]| token, used for classification. We use the original TensorFlow \citep{Abadi_Tensorflow} implementation of BERT\footnote{\url{https://github.com/google-research/bert}} and focus on the 12-layer \texttt{bert\_base\_uncased} variant. We denote the pre-trained model as \textbf{Base} and refer to fine-tuned versions by the name of the task.

\paragraph{MNLI} A common benchmark for natural language understanding, the MNLI dataset \cite{Williams_MNLI} contains 
over 433K sentence pairs annotated with textual entailment information. BERT is adapted to this task by feeding the \verb|[CLS]| token in the last layer through an additional output layer for predictions. We use the parameters and architecture of \citet{Devlin_BERT} for fine-tuning. Across three trials, the evaluation accuracy of our BERT Base model is $83.3 \pm 0.1$, slightly lower but comparable to the published score of $84.6$. 

\paragraph{SQuAD} The SQuADv1.1 dataset \cite{Rajpurkar_SQuAD} contains over 100,000 crowd-sourced question-answer pairs, created from a set of Wikipedia articles. The answers are given by contiguous spans from the original question, so the task is integrated into the BERT framework with an additional output layer for predicting the start and end tokens of the answer. We use the parameters and architecture of \citet{Devlin_BERT} for fine-tuning. Our average F1 score is $89.2 \pm 0.2$, slightly higher than the published $88.5$. 

\paragraph{Dependency Parsing} We also introduce a BERT model fine-tuned on dependency parsing (Dep). We include this task to present a contrasting perspective from the prior two datasets, in particular since prior research has suggested that much of the information needed to solve dependency parsing is already present in the pre-trained base \cite{Hewitt_A_Structural, Goldberg_Assessing}. Our model is trained on data from the CoNLL 2017 Shared Task \cite{Zeman_CoNLL} and uses the features of BERT as input to a biaffine classifier, similar to \citet{Dozat_Deep}. The model uses a learning rate of $3 \times 10^{-5}$ with a $10\%$ warm-up portion, uses an Adam optimizer \cite{Kingma_Adam}, and is trained for 20 epochs. The Labeled Attachment Score (LAS) on the development set for our models is $96.3 \pm 0.1$.

\section{What happens to linguistic features?}
\label{sec:probing}

Equipped with the models trained on these downstream tasks, we ask how the representation of linguistic features in the fine-tuned models compare to those in the pre-trained model? Recent studies have shown that these robust features do not inform predictions on downstream tasks, with models appearing to use dataset heuristics such as lexical overlap \cite{Mccoy_Right} or word priors \citep{Poliak_Hypothesis}, but it is an open question whether this is because these features are forgotten entirely or simply are not always used. We approach this with supervised probing models, using two complementary techniques: edge probes \citep{Tenney_What} which test for labeling information across several formalisms, and structural probes \citep{Hewitt_A_Structural} which measure the representation of syntactic structure.

\begin{table*}
    \centering
    \begin{tabular}{l||c||cc||ccc}
        \multicolumn{2}{c}{} & \multicolumn{2}{c}{$\Delta$ \textbf{for Baselines}} & \multicolumn{3}{c}{$\Delta$ \textbf{for Fine-tuned Models}} \\ 
         \hline
         \textbf{Task} & \textbf{BERT Base} & \textbf{Lexical} & \textbf{Randomized} & \textbf{MNLI} & \textbf{SQuAD}  & \textbf{Dep} \\
         POS & 97.53 & -8.99 & -13.61 & -0.17 & \textbf{-1.52} & -0.22 \\
         Constituents & 84.35 & -24.13 & -12.88 & \textbf{-2.18} & 0.08 & \textbf{4.35} \\
         Dependencies & 95.45  & -15.57 & -18.19 & -0.51 & \textbf{-2.49} & 0.18 \\
         Entities & 96.19 & -6.56 & -9.97 & -0.34 & -0.96 & -0.62 \\
         SRL & 92.87  & -13.57 & -15.04 & -0.36 & \textbf{-2.90 }& -0.50 \\
         Coreference & 95.72 & -5.82 & -6.15 &  -0.49 & -0.84 & \textbf{-1.22} \\
         SPR & 84.61 & -6.56 & -12.21 & -0.67 & -0.40 & \textbf{-1.17}  \\ 
         Relations & 79.50 & -20.68 & -40.53 & -0.75 & -0.37 &\textbf{ -2.53}
    \end{tabular}
    \caption{Comparison of F1 performance on the edge probing tasks before and after fine-tuning. The BERT Base performance is consistent with \cite{Tenney_What}, and the results show that the fine-tuned models retain most of the linguistic concepts discovered during unsupervised pre-training. We report single numbers for clarity, but note that variation across runs is $\pm0.5$ between probing runs, $\pm0.7$ between fine-tuning runs from the same checkpoint, and $\pm1.0$ point between different pretraining runs.
    }
    \label{fig:mnli_edge}
\end{table*}

\paragraph{Notation}
Let $m$ be the hidden size of the attention layers of a Transformer. Then, we define $\bm{h}_i^l \in \mathbb{R}^m$ to be the hidden representation of the $i$-th word at the $l$-th attention layer. Note that BERT uses subword tokenization, so word representations are aggregated using mean pooling over all subword components.

\paragraph{Edge Probing}
The edge probing tasks of \citet{Tenney_What} aim to measure how well a contextual encoder captures linguistic phenomena, ranging from syntactic concepts such as part-of-speech to more semantic abstractions including entity typing and relation classification. To perform this task, the trained encoder is frozen, and the relevant hidden states $\bm{h}_i^l$ are fed into an auxiliary shallow neural network to predict labeling information. We use the tasks, architecture, and procedure of \citet{Tenney_What}.\footnote{The dependency labeling task is from the English Web Treebank \cite{Silveira_A}, SPR corresponds to SPR1 from \citet{Teichert_Semantic}, and relations is Task 8 from SemEval 2010 \cite{Hendrickx_Semeval}. All of the other tasks are from OntoNotes 5.0 \cite{Weischedel_Ontonotes}.} After training, we report the micro-averaged F1 scores on a held-out test set.

\paragraph{Structural Probe}
The structural probes of \citet{Hewitt_A_Structural} also analyze the token representations but are designed to evaluate how well this space encodes syntactic structure. Specifically, the probe identifies whether the squared L2 distance of the representations under some linear transformation encodes the tree distances between words in the dependency parse. The initial paper's results showed that the deep models (ELMo and BERT) discover this syntax information during pre-training, which is not present in simple word embedding baselines.

The first structural probe predicts the tree depth (tree distance from the root node) by: $$||\bm{h}_i^l||_B = (B \bm{h}_i^l)^\top (B \bm{h}_i^l)$$ where $B \in \mathbb{R}^{k \times m}$ is a learned matrix.\footnote{For our experiments, we use a rank $k=512$ to match the projection dimension from the edge probes.} The corresponding metrics are the prediction accuracy of the root node and the Spearman correlation between the predicted and true tree depths.

The second structural probe measures the pairwise distances for all words in the parse tree. For any two words ($\bm{h}_i^l$, $\bm{h}_j^l$), the distance is given by: $$||\bm{h}_i^l - \bm{h}_j^l||_B = (B(\bm{h}_i^l - \bm{h}_j^l))^\top (B(\bm{h}_i^l - \bm{h}_j^l))$$ Following \citet{Hewitt_A_Structural}, we evaluate Spearman correlation between each row  of the predicted and true distance matricies, and also compute the minimum spanning tree and compare to the true parse using the Undirected Unlabeled Attachment Score (UUAS).

\begin{figure*}
    \centering
    
    \begin{subfigure}{0.9\textwidth}
            \begin{subfigure}{0.5\textwidth}
                \centering
                \includegraphics[width=\textwidth, trim=4 4 4 4, clip]{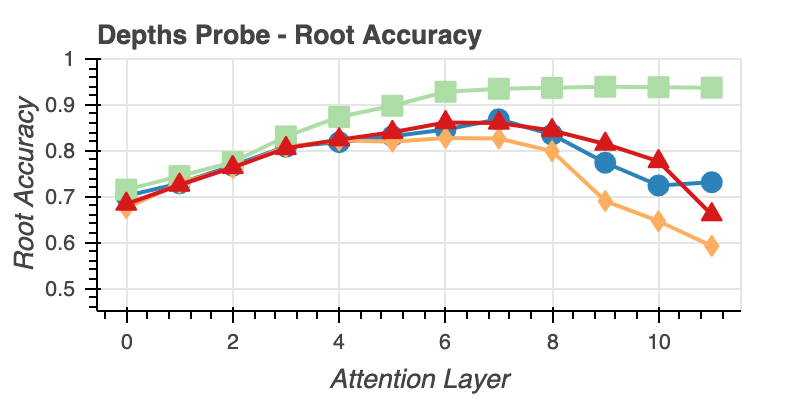}
            \end{subfigure}%
            \begin{subfigure}{0.5\textwidth}
                \centering
                \includegraphics[width=\textwidth, trim=4 4 4 4, clip]{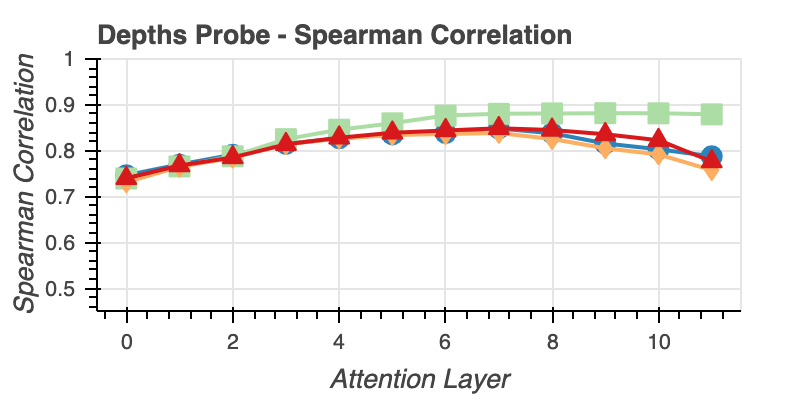}
            \end{subfigure}
            
            \begin{subfigure}{0.5\textwidth}
                \centering
                \includegraphics[width=\textwidth, trim=4 4 4 4, clip]{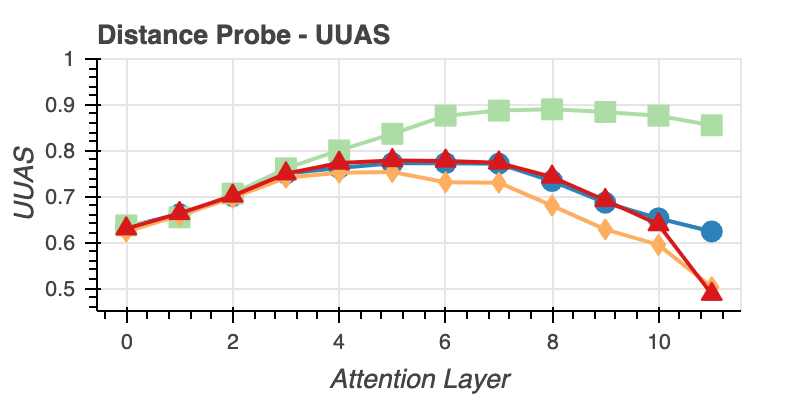}
            \end{subfigure}%
            \begin{subfigure}{0.5\textwidth}
                \centering
                \includegraphics[width=\textwidth, trim=4 4 4 4, clip]{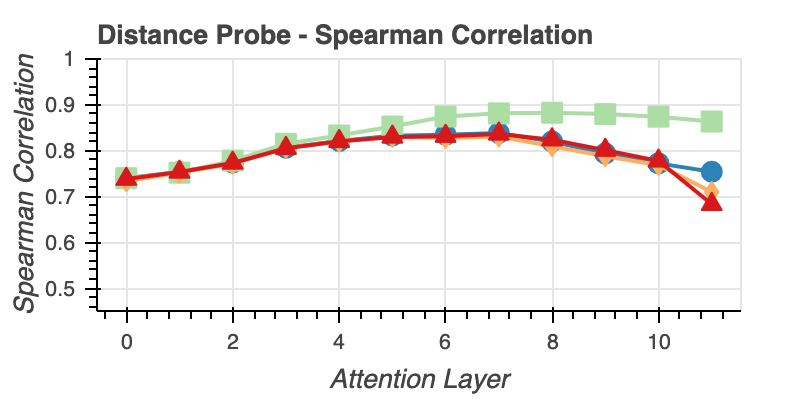}
            \end{subfigure}
    \end{subfigure}%
    \begin{subfigure}{0.1\textwidth}
            \begin{subfigure}{\textwidth}
                \centering
                \includegraphics[width=\textwidth]{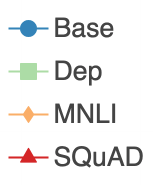}
            \end{subfigure}
    \end{subfigure}
    
    \caption{Comparison of the structural probe performance on BERT models before and after fine-tuning. The stability of the Spearman correlations between both the depths and distance probes suggest that the embeddings still retain significant information about the syntax of inputted sentences.
    }
    \label{fig:mnli_structural}
\end{figure*}

\subsection{Results}

The results from both probing tasks largely demonstrate that the linguistic structures from pre-training are preserved in the fine-tuned models. This is first seen in the edge probing metrics presented in Table~\ref{fig:mnli_edge}. For the sake of comparison, we also provide baseline results on the output of the embedding layer (Lexical) and a randomly initialized BERT architecture (Randomized). These baselines are important because inspection-based analysis can often discover patterns that are not obviously present due to the high capacity of auxiliary classifiers. For example, \citet{Zhang_Language, Hewitt_Designing} found that expressive-enough probing methods can have decent performance even when trained on random noise. 

Across the edge probing suite, for all three tasks we see only small changes in F1 score compared to BERT base. In most cases we observe a drop in performance of 0.5-2\%, with some variation: MNLI and SQuAD lead to drops of 1.5-3\% on syntactic tasks--constituents, and POS, dependencies, and SRL, respectively--while the dependency parsing model leads to significantly improved syntactic performance (+4\% on constituent labeling) while dropping performance on the more semantically-oriented coreference, SPR, and relation classification tasks. Nonetheless, in all cases these effects are small: they are comparable to the variation between randomly-seeded fine-tuning runs ($\pm0.7$), and much smaller than the difference between the full model and the Lexical or Randomized baselines, suggesting that most linguistic information from BERT is still available after fine-tuning.

Next, we turn to the structural probe, with results seen in Figure \ref{fig:mnli_structural}. First, the dependency parsing fine-tuned model shows improvements in both the correlation and the absolute metrics of root accuracy and UUAS, as early as layer 5. Since the structural probes are designed and trained to look for syntax, this result suggests that the fine-tuning improves the model's internal representation of such information.
This makes intuitive sense as the fine-tuning task is aligned with the probing task. 

On the MNLI and SQuAD fine-tuned models, we observe small drops in performance, particularly with the final layer. These changes are most pronounced on the root accuracy and UUAS metrics, which score against a discrete decoded solution (\verb|argmin| for root accuracy or minimum spanning tree for UUAS), but are smaller in magnitude on Spearman correlations which consider all predictions. This suggests that while some information is lost, the actual magnitude of change within the ``syntactic subspace" is quite small. This is consistent with observations by \citet{Gauthier_Linking} and suggests that information about syntactic structure is well-preserved in end-task models.

Overall, the results from these two probing techniques suggest that there is no catastrophic forgetting. This is surprising as a number of prior error analyses have shown that the fine-tuned models often do not use syntax \cite{Mccoy_Right} and rely on annotation artifacts \cite{Gururangan_Annotation} or simple pattern matching \citep{Jia_Adversarial} to solve downstream tasks. 
Our analysis suggests that the while this linguistic information may not be incorporated into the final predictions, it is still available in the model's representations.

\section{What changes in the representations?}
\label{sec:rep_change}
Supervised probes are highly targeted: as trained models, they are sensitive to particular linguistic phenomena, but they also can learn to ignore everything else. If the supervised probe is closely related to the fine-tuning task--such as for syntactic probes and a dependency parsing model--we have observed significant changes in performance, but otherwise we see little effect. Nonetheless, we know that \textit{something} must be changing when fine-tuning--as evidenced by prior work that shows that end-task performance degrades if the encoder is completely frozen \cite{Peters_Tune}. To explore this change more broadly, we turn to an unsupervised technique, Representational Similarity Analysis, and corroborate our findings with layer-based ablations.

\subsection{Representational Similarity Analysis}
\label{sec:rsa}
Representational Similarity Analysis \citep[RSA; ][]{Laakso_Content} is a technique for measuring the similarity between two different representation spaces for a given set of stimuli. Originally developed for neuroscience \cite{Kriegeskorte_Representational}, it has become increasingly used to analyze similarity between neural network activations \citep{Abnar_Blackbox,Chrupala_Correlating}. The method works by using a common set of $n$ examples, used to create two comparable sets of representations. Using two kernels (possibly different for each representation space) to measure the similarity of paired examples, the sets of representations are then converted into two pairwise similarity matrices in $\mathbb{R}^{n \times n}$. The final similarity score between the two representation spaces is calculated as the Pearson correlation between the flattened upper triangulars of the two similarity matrices. %

In our application, we pass ordinary sentences (Wikipedia), sentence-pairs (MNLI), or question-answer pairs (SQuAD) as inputs to the BERT model, and select a random subset ($n = 5000$) of tokens as our stimuli. This choice of input is consistent with the masked language model pre-training and various fine-tuning tasks (such as SQuAD and dependency parsing) in analyzing the contextual representations for every token. We extract representations as the activations of corresponding layers from the two models to compare (e.g. Base vs. a fine-tuned model). Following previous applications of RSA to text representations \citep{Abnar_Blackbox, Chrupala_Correlating}, we adopt cosine similarity as the kernel for all of our experiments.

While RSA does not require learning any parameters and is thus resistant to overfitting \cite{Abdou_Higher}, the unsupervised technique is sensitive to spurious signals in the representations that may not be relevant to model behavior.\footnote{We note that probing techniques are more robust to this, since they learn to focus on relevant features.}
 To mitigate this, we repeat the BERT pre-training procedure (as described in Section 3 of \citep{Devlin_BERT}) from scratch three times. For each pre-trained checkpoints, we fine-tune on the three downstream task and report the average for these independent runs.

\begin{figure}
    \centering
    \begin{subfigure}{0.45\textwidth}
        \includegraphics[width=\textwidth, trim=4 4 4 4, clip]{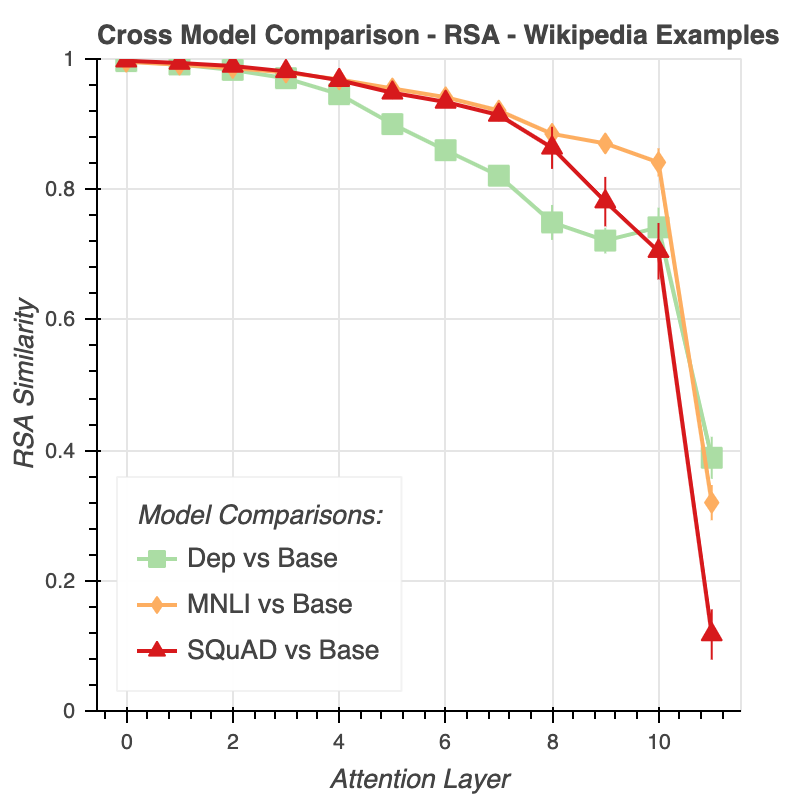}
    \end{subfigure}
    \caption{Comparison of the representations from BERT base and the various fine-tuned models, when tested on Wikipedia examples. The dependency probing model starts to diverge from BERT Base around layer 5, which matches the previous results from edge probing. For the MNLI and SQuAD fine-tuned models, the differences from the Base model mainly arise in the top layers of the network. 
    }
    \label{fig:dep_rsa}
\end{figure}

\paragraph{Results} 

Figure \ref{fig:dep_rsa} shows the results of our RSA analysis comparing the three task models, \textbf{Dep.}, \textbf{MNLI}, and \textbf{SQuAD}, to BERT \textbf{Base} at each layer, using single-sentence inputs randomly selected from English Wikipedia. Across all tasks, we observe that changes generally arise in the top layers of the network, with very little change observed in the layers closest to the input. To first order, this may be a result of optimization: vanishing gradients result in the most change in the layers closest to the loss. Yet we do observe significant differences between tasks. Except for the output layer which is particularly sensitive to the form of the output (span-based for dependencies and SQuAD, or using the \verb|[CLS]| token for MNLI), we see that MNLI involves the smallest changes to the model: the second-to-last attention layer still shows a very high similarity score of $0.84 \pm 0.02$ compared to the representations of the pretrained encoder. The SQuAD model shows a steeper change, behaving similarly to the Base model through layer 7 but dropping off steeply afterwards - suggesting that fine-tuning on this task involves a deeper, yet still relatively shallow reconfiguration of the encoder.

Finally, dependency parsing presents a dissimilar pattern: we observe the deepest changes, departing from the Base model as early as layers 4 and 5. This is consistent with our supervised probing observations (Section~\ref{sec:probing}), in which we observe improved performance on syntactic features at a similar point in the model (Figure~\ref{fig:mnli_structural}).

\subsection{Layer Ablations}
\label{sec:ablations}

\begin{figure}[t]
    \centering
    
    \begin{subfigure}{0.45\textwidth}
        \centering
        \includegraphics[width=\textwidth, trim=4 4 4 4, clip]{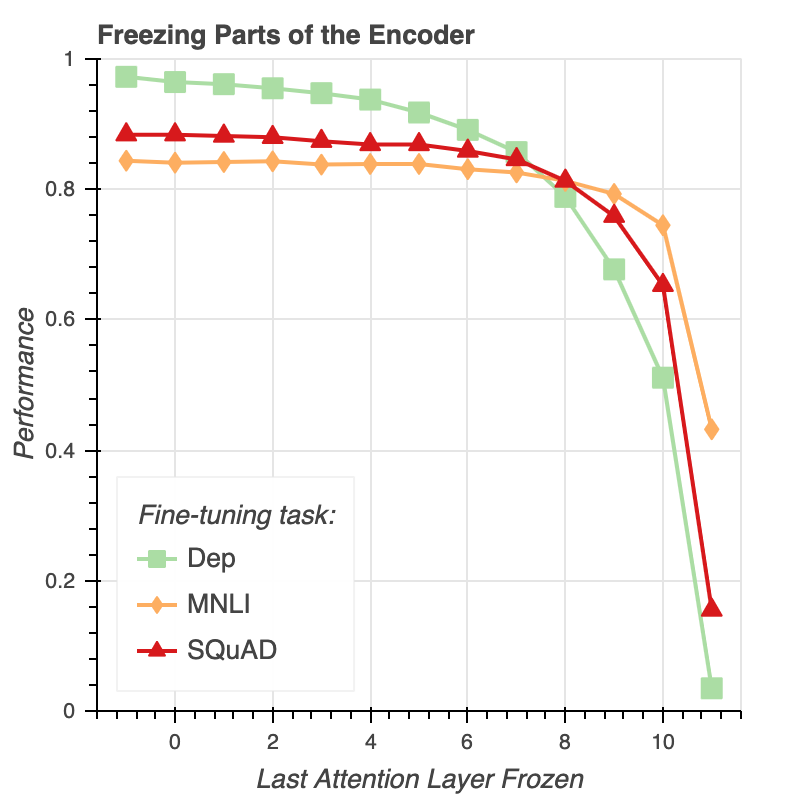}
    \end{subfigure}
    
    \caption{Effects of freezing an increasing number of layers during fine-tuning on performance, where different lines correspond to various tasks (we report the evaluation accuracy for MNLI, F1 score for SQuAD, and LAS for Dep).
    The point at -1 corresponds to no frozen components. The graph shows that only a few unfrozen layers are needed to improve task performance, supporting the shallow processing conclusion.}
    \label{fig:partial_freezing}
\end{figure}

\begin{figure}[t]
    \centering
    
    \begin{subfigure}{0.45\textwidth}
        \centering
        \includegraphics[width=\textwidth, trim=4 4 4 4, clip]{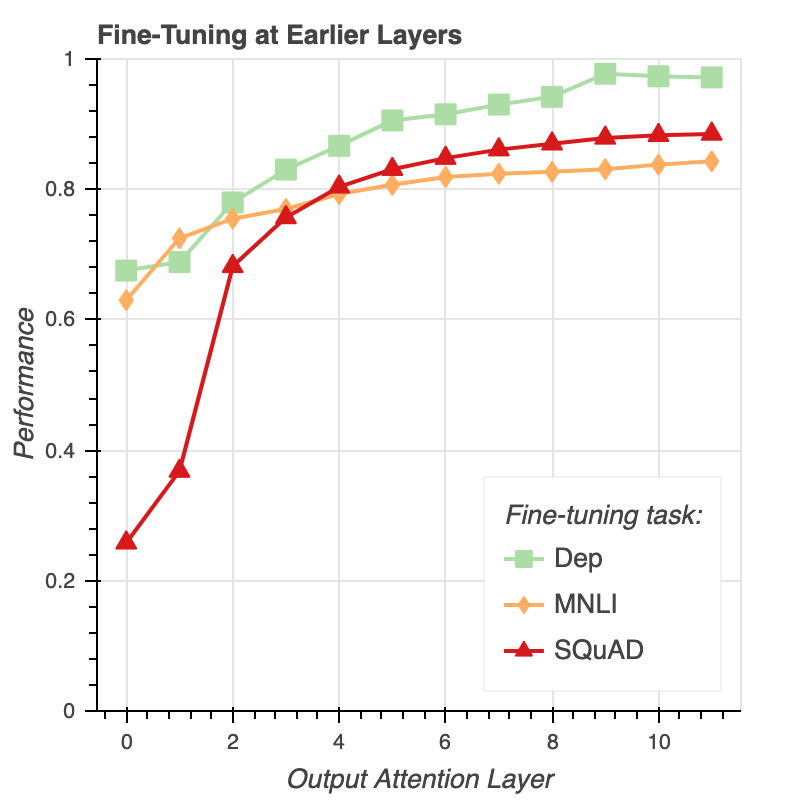}
    \end{subfigure}
    
    \caption{Effects of fine-tuning at earlier layers of BERT. We note that the MNLI evaluation accuracy and SQuAD F1 score approach the full model performance by layer 6, whereas the dependency parsing LAS seems to require more layers. This support our hypothesis that the processing of dataset-specific features is shallow.}
    \label{fig:early_layer}
\end{figure}

As an unsupervised, metric-based technique, RSA tells us about broad changes in the representation space, but does not in itself say if these changes are important for the model's behavior--i.e. for the processing necessary to solve the downstream task. To measure our observations in terms of task performance, we turn to two layer ablation studies.

\textbf{Partial Freezing} can be thought of as a test for how many layers \textit{need} to change for a downstream task. We freeze the bottom $k$ layers (and the embeddings)--treating them as features--but allow the rest to adapt. Effectively, this clamps the first $k$ layers to have RSA similarity of 1 with the Base model. Also, we perform \textbf{model truncation} as a rough estimate of difficulty for each task, and as an attempt to de-couple the results of partial freezing from helpful features that may be available in top layers of BERT Base \citep{Tenney_BERT}. Figure~\ref{fig:partial_freezing} (partial freezing) Figure~\ref{fig:early_layer} (truncation) show the effect on task performance.

The patterns we observe corroborate the findings of our RSA analysis. On MNLI, we find that performance does not drop significantly unless the last two layers are frozen, while the truncated models are able to achieve comparable performance with only three attention layers. This suggests that while natural language inference \cite{Dagan_The} is known to be a complex task \textit{in the limit}, most MNLI examples can be resolved with relatively shallow processing. SQuAD exhibits a similar trend: we see a significant performance drop when 3 or fewer layers are allowed to change (e.g. freezing through layer 8 or higher), consistent with where RSA finds the greatest change. From our truncation experiment, we similarly see that only five layers are needed to achieve comparable performance to the full model. 

\begin{figure*}[ht!]
    \centering
    \begin{subfigure}{0.5\textwidth}
        \centering
        \includegraphics[width=0.9\textwidth, trim=4 4 4 4, clip]{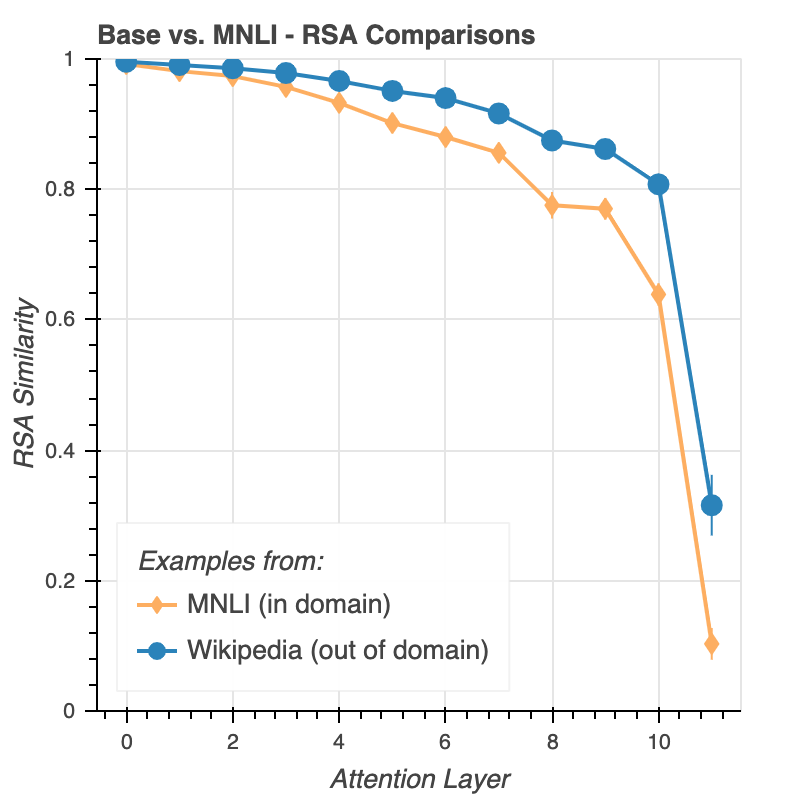}
    \end{subfigure}%
    \begin{subfigure}{0.5\textwidth}
        \centering
        \includegraphics[width=0.9\textwidth, trim=4 4 4 4, clip]{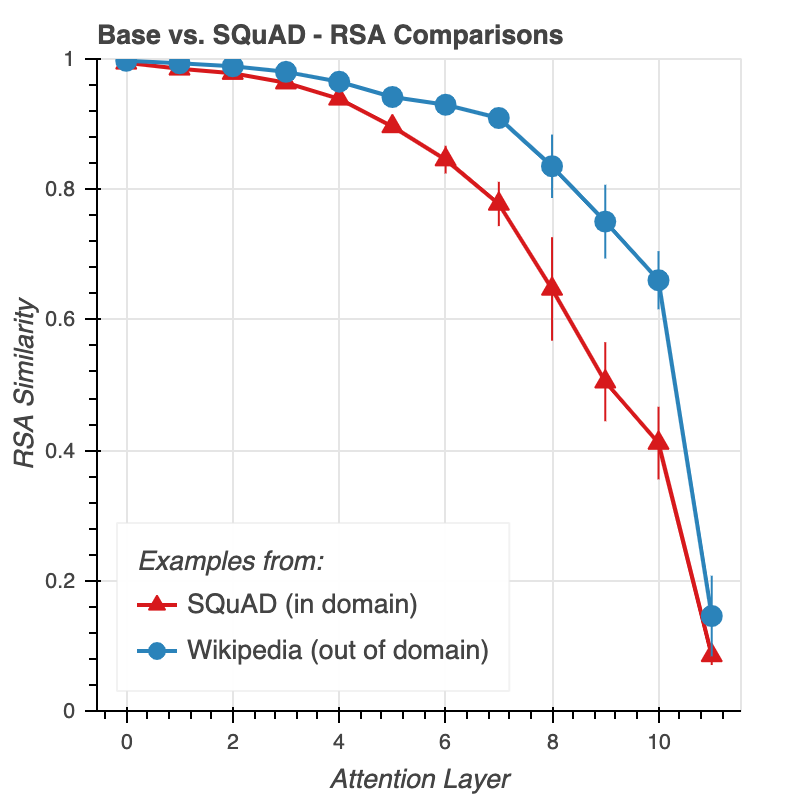}
    \end{subfigure}
    \caption{Comparison of the representations in the MNLI (left) and SQuAD (right) fine-tuned models and those of BERT Base, with the different lines corresponding to examples coming from various datasets. These graphs show that fine-tuning models only lead to shallow changes, consolidated to the last few layers. Also, we see that fine-tuning has a much greater impact on the token representations of in-domain data.}
    \label{fig:rsa_domain}
\end{figure*}

Dependency parsing performance drops more rapidly still--in both experiments--again consistent with the previous results from RSA. This is perhaps surprising, since probing analysis \cite{Goldberg_Assessing, Marvin_Targeted} suggests that many syntactic phenomena are well-captured by the pre-trained model, and diagnostics for dependency parsing in particular \citep{Tenney_What,Tenney_BERT,Hewitt_A_Structural, Clark_What} show strong performance from probes on frozen models. Yet as observed with the structural probes (Figure~\ref{fig:mnli_structural}) there is much headroom available, and it appears that to capture it requires changing deeper parts of the model. We hypothesize that this effect may come from the hierarchical nature of parsing, which requires additional layers to determine the tree-like structure, and fully reconciling these observations would be a promising direction for future work.

\section{Out-of-Domain Behavior}
\label{sec:generalization}

Finally, we ask whether the effects of fine-tuning are general: do they apply only to in-domain inputs, or do they lead to broader changes in behavior? This is usually explored by behavioral methods, in which a model is trained on one domain and evaluated on another--for example, the mismatched evaluation for MNLI \citep{Williams_MNLI}--but this analysis is limited by the availability of labeled data. By using Representational Similarity Analysis, we can test this in an unsupervised manner.

We use RSA to compare the fine-tuned model to Base and observe the degree of similarity when inputs are drawn from different corpora. We use random samples from the development sets for MNLI (as \texttt{premise [SEP] hypothesis}) and SQuAD (as \texttt{question [SEP] passage}) as in-domain for their respective models,\footnote{Note that these are unseen during fine-tuning, although RSA scores do not change significantly if the MNLI or SQuAD training sets are used.} and as the out-of-domain control we use random Wikipedia sentences (which resemble the pre-training domain).
As in Section~\ref{sec:rsa}, we use the representations of $n=5000$ tokens as our stimuli for each comparison.\footnote{We also tested single-sentence examples from MNLI and SQuAD by only taking the premise and question respectively; the trends were similar to Figure~\ref{fig:rsa_domain}.
 }
 Results for the MNLI and SQuAD fine-tuned models are shown in Figure~\ref{fig:rsa_domain}.

We note that in both cases, the trend follows Section~\ref{sec:rsa} in that the models diverge from BERT Base in the top layers, but there is a significantly larger change in representations on the in-domain examples. When evaluated on Wikipedia sentences, which resemble the pre-training data, the similarity scores are much higher. This suggests that fine-tuning leads the model to change its representations for the fine-tuning domain but to continue to behave more like the Base model otherwise.

\section{Conclusions}

In this paper, we employ three complementary analysis methods to gain insight on the effects of fine-tuning on the representations produced by BERT. From supervised probing analyses, we find that the linguistic structures discovered during pre-training remain available after fine-tuning. Prior studies \citep{Mccoy_Right,Jia_Adversarial} have shown that end-task models often fall back on simple heuristics; taken together, our results suggest that while present, perhaps the linguistic features are not always incorporated into final predictions.

Next, our results using RSA and layer ablations show that the changes from fine-tuning alter a fraction of the model capacity, specifically within the top few layers (up to some variation across tasks). Also, although fine-tuning has a significant affect on the representations of in-domain sentences, the representations of out-of-domain examples remain much closer to those of the pre-trained model.

Overall, these conclusions suggest that fine-tuning--as currently practiced--is a conservative process: largely preserving linguistic features, affecting only a few layers, and specific to in-domain examples. While the standard fine-tuning recipe undeniably leads to strong performance on many tasks, there appears to be room for improvement: an opportunity to refine this transfer step--potentially by utilizing more of the model capacity--to better the generalization and transferability.

\bibliography{anthology,main}

\begin{thebibliography}{64}
\expandafter\ifx\csname natexlab\endcsname\relax\def\natexlab#1{#1}\fi

\bibitem[{Abadi et~al.(2015)Abadi, Agarwal, Barham, Brevdo, Chen, Citro,
  Corrado, Davis, Dean, Devin, Ghemawat, Goodfellow, Harp, Irving, Isard, Jia,
  Jozefowicz, Kaiser, Kudlur, Levenberg, Man\'{e}, Monga, Moore, Murray, Olah,
  Schuster, Shlens, Steiner, Sutskever, Talwar, Tucker, Vanhoucke, Vasudevan,
  Vi\'{e}gas, Vinyals, Warden, Wattenberg, Wicke, Yu, and
  Zheng}]{Abadi_Tensorflow}
Mart\'{\i}n Abadi, Ashish Agarwal, Paul Barham, Eugene Brevdo, Zhifeng Chen,
  Craig Citro, Greg~S. Corrado, Andy Davis, Jeffrey Dean, Matthieu Devin,
  Sanjay Ghemawat, Ian Goodfellow, Andrew Harp, Geoffrey Irving, Michael Isard,
  Yangqing Jia, Rafal Jozefowicz, Lukasz Kaiser, Manjunath Kudlur, Josh
  Levenberg, Dandelion Man\'{e}, Rajat Monga, Sherry Moore, Derek Murray, Chris
  Olah, Mike Schuster, Jonathon Shlens, Benoit Steiner, Ilya Sutskever, Kunal
  Talwar, Paul Tucker, Vincent Vanhoucke, Vijay Vasudevan, Fernanda Vi\'{e}gas,
  Oriol Vinyals, Pete Warden, Martin Wattenberg, Martin Wicke, Yuan Yu, and
  Xiaoqiang Zheng. 2015.
\newblock \href {https://www.tensorflow.org/} {{TensorFlow}: Large-scale
  machine learning on heterogeneous systems}.
\newblock Software available from tensorflow.org.

\bibitem[{Abdou et~al.(2019)Abdou, Kulmizev, Hill, Low, and
  S{\o}gaard}]{Abdou_Higher}
Mostafa Abdou, Artur Kulmizev, Felix Hill, Daniel~M. Low, and Anders
  S{\o}gaard. 2019.
\newblock \href {https://doi.org/10.18653/v1/D19-1593} {Higher-order
  comparisons of sentence encoder representations}.
\newblock In \emph{Proceedings of the 2019 Conference on Empirical Methods in
  Natural Language Processing and the 9th International Joint Conference on
  Natural Language Processing (EMNLP-IJCNLP)}, pages 5837--5844, Hong Kong,
  China. Association for Computational Linguistics.

\bibitem[{Abnar et~al.(2019)Abnar, Beinborn, Choenni, and
  Zuidema}]{Abnar_Blackbox}
Samira Abnar, Lisa Beinborn, Rochelle Choenni, and Willem Zuidema. 2019.
\newblock \href {https://doi.org/10.18653/v1/W19-4820} {Blackbox meets
  blackbox: Representational similarity {\&} stability analysis of neural
  language models and brains}.
\newblock In \emph{Proceedings of the 2019 ACL Workshop BlackboxNLP: Analyzing
  and Interpreting Neural Networks for NLP}, pages 191--203, Florence, Italy.
  Association for Computational Linguistics.

\bibitem[{Adi et~al.(2016)Adi, Kermany, Belinkov, Lavi, and
  Goldberg}]{Adi_Fine}
Yossi Adi, Einat Kermany, Yonatan Belinkov, Ofer Lavi, and Yoav Goldberg. 2016.
\newblock Fine-grained analysis of sentence embeddings using auxiliary
  prediction tasks.
\newblock \emph{CoRR}, abs/1608.04207.

\bibitem[{van Aken et~al.(2019)van Aken, Winter, Löser, and Gers}]{Aken_How}
Betty van Aken, Benjamin Winter, Alexander Löser, and Felix~A. Gers. 2019.
\newblock \href {http://arxiv.org/abs/1909.04925} {How does bert answer
  questions? a layer-wise analysis of transformer representations}.

\bibitem[{Belinkov and Glass(2019)}]{Belinkov_Analysis}
Yonatan Belinkov and James Glass. 2019.
\newblock \href {https://doi.org/10.1162/tacl_a_00254} {Analysis methods in
  neural language processing: A survey}.
\newblock \emph{Transactions of the Association for Computational Linguistics},
  7:49--72.

\bibitem[{Blevins et~al.(2018)Blevins, Levy, and Zettlemoyer}]{Blevins_Deep}
Terra Blevins, Omer Levy, and Luke Zettlemoyer. 2018.
\newblock \href {https://doi.org/10.18653/v1/P18-2003} {Deep {RNN}s encode soft
  hierarchical syntax}.
\newblock In \emph{Proceedings of the 56th Annual Meeting of the Association
  for Computational Linguistics (Volume 2: Short Papers)}, pages 14--19,
  Melbourne, Australia. Association for Computational Linguistics.

\bibitem[{Brunner et~al.(2019)Brunner, Liu, Pascual, Richter, Ciaramita, and
  Wattenhofer}]{Brunner_On}
Gino Brunner, Yang Liu, Damián Pascual, Oliver Richter, Massimiliano
  Ciaramita, and Roger Wattenhofer. 2019.
\newblock \href {http://arxiv.org/abs/1908.04211} {On identifiability in
  transformers}.

\bibitem[{Chrupa{\l}a and Alishahi(2019)}]{Chrupala_Correlating}
Grzegorz Chrupa{\l}a and Afra Alishahi. 2019.
\newblock \href {https://doi.org/10.18653/v1/P19-1283} {Correlating neural and
  symbolic representations of language}.
\newblock In \emph{Proceedings of the 57th Annual Meeting of the Association
  for Computational Linguistics}, pages 2952--2962, Florence, Italy.
  Association for Computational Linguistics.

\bibitem[{Clark et~al.(2019)Clark, Khandelwal, Levy, and Manning}]{Clark_What}
Kevin Clark, Urvashi Khandelwal, Omer Levy, and Christopher~D. Manning. 2019.
\newblock \href {https://doi.org/10.18653/v1/W19-4828} {What does {BERT} look
  at? an analysis of {BERT}{'}s attention}.
\newblock In \emph{Proceedings of the 2019 ACL Workshop BlackboxNLP: Analyzing
  and Interpreting Neural Networks for NLP}, pages 276--286, Florence, Italy.
  Association for Computational Linguistics.

\bibitem[{Coenen et~al.(2019)Coenen, Reif, Yuan, Kim, Pearce, Vi{\'{e}}gas, and
  Wattenberg}]{Coenen_Visualizing}
Andy Coenen, Emily Reif, Ann Yuan, Been Kim, Adam Pearce, Fernanda~B.
  Vi{\'{e}}gas, and Martin Wattenberg. 2019.
\newblock Visualizing and measuring the geometry of {BERT}.
\newblock \emph{CoRR}, abs/1906.02715.

\bibitem[{Conneau et~al.(2018)Conneau, Kruszewski, Lample, Barrault, and
  Baroni}]{Conneau_What}
Alexis Conneau, German Kruszewski, Guillaume Lample, Lo{\"\i}c Barrault, and
  Marco Baroni. 2018.
\newblock \href {https://doi.org/10.18653/v1/P18-1198} {What you can cram into
  a single {\$}{\&}!{\#}* vector: Probing sentence embeddings for linguistic
  properties}.
\newblock In \emph{Proceedings of the 56th Annual Meeting of the Association
  for Computational Linguistics (Volume 1: Long Papers)}, pages 2126--2136,
  Melbourne, Australia. Association for Computational Linguistics.

\bibitem[{Dagan et~al.(2006)Dagan, Glickman, and Magnini}]{Dagan_The}
Ido Dagan, Oren Glickman, and Bernardo Magnini. 2006.
\newblock \href {https://doi.org/10.1007/11736790_9} {The pascal recognising
  textual entailment challenge}.
\newblock In \emph{Proceedings of the First International Conference on Machine
  Learning Challenges: Evaluating Predictive Uncertainty Visual Object
  Classification, and Recognizing Textual Entailment}, MLCW'05, pages 177--190,
  Berlin, Heidelberg. Springer-Verlag.

\bibitem[{Devlin et~al.(2019)Devlin, Chang, Lee, and Toutanova}]{Devlin_BERT}
Jacob Devlin, Ming-Wei Chang, Kenton Lee, and Kristina Toutanova. 2019.
\newblock \href {https://doi.org/10.18653/v1/N19-1423} {{BERT}: Pre-training of
  deep bidirectional transformers for language understanding}.
\newblock In \emph{Proceedings of the 2019 Conference of the North {A}merican
  Chapter of the Association for Computational Linguistics: Human Language
  Technologies, Volume 1 (Long and Short Papers)}, pages 4171--4186,
  Minneapolis, Minnesota. Association for Computational Linguistics.

\bibitem[{Dozat and Manning(2017)}]{Dozat_Deep}
Timothy Dozat and Christopher~D. Manning. 2017.
\newblock Deep biaffine attention for neural dependency parsing.
\newblock In \emph{{ICLR} (Poster)}. OpenReview.net.

\bibitem[{Ettinger(2020)}]{Ettinger_BertNot}
Allyson Ettinger. 2020.
\newblock What bert is not: Lessons from a new suite of psycholinguistic
  diagnostics for language models.
\newblock \emph{Transactions of the Association for Computational Linguistics},
  8:34--48.

\bibitem[{Ettinger et~al.(2018)Ettinger, Elgohary, Phillips, and
  Resnik}]{Ettinger_Assessing}
Allyson Ettinger, Ahmed Elgohary, Colin Phillips, and Philip Resnik. 2018.
\newblock \href {https://www.aclweb.org/anthology/C18-1152} {Assessing
  composition in sentence vector representations}.
\newblock In \emph{Proceedings of the 27th International Conference on
  Computational Linguistics}, pages 1790--1801, Santa Fe, New Mexico, USA.
  Association for Computational Linguistics.

\bibitem[{Gauthier and Levy(2019)}]{Gauthier_Linking}
Jon Gauthier and Roger Levy. 2019.
\newblock \href {https://doi.org/10.18653/v1/D19-1050} {Linking artificial and
  human neural representations of language}.
\newblock In \emph{Proceedings of the 2019 Conference on Empirical Methods in
  Natural Language Processing and the 9th International Joint Conference on
  Natural Language Processing (EMNLP-IJCNLP)}, pages 529--539, Hong Kong,
  China. Association for Computational Linguistics.

\bibitem[{Goldberg(2019)}]{Goldberg_Assessing}
Yoav Goldberg. 2019.
\newblock Assessing bert's syntactic abilities.
\newblock \emph{CoRR}, abs/1901.05287.

\bibitem[{Gulordava et~al.(2018)Gulordava, Bojanowski, Grave, Linzen, and
  Baroni}]{Gulordava_Colorless}
Kristina Gulordava, Piotr Bojanowski, Edouard Grave, Tal Linzen, and Marco
  Baroni. 2018.
\newblock \href {https://doi.org/10.18653/v1/N18-1108} {Colorless green
  recurrent networks dream hierarchically}.
\newblock In \emph{Proceedings of the 2018 Conference of the North {A}merican
  Chapter of the Association for Computational Linguistics: Human Language
  Technologies, Volume 1 (Long Papers)}, pages 1195--1205, New Orleans,
  Louisiana. Association for Computational Linguistics.

\bibitem[{Gururangan et~al.(2018)Gururangan, Swayamdipta, Levy, Schwartz,
  Bowman, and Smith}]{Gururangan_Annotation}
Suchin Gururangan, Swabha Swayamdipta, Omer Levy, Roy Schwartz, Samuel Bowman,
  and Noah~A. Smith. 2018.
\newblock \href {https://doi.org/10.18653/v1/N18-2017} {Annotation artifacts in
  natural language inference data}.
\newblock In \emph{Proceedings of the 2018 Conference of the North {A}merican
  Chapter of the Association for Computational Linguistics: Human Language
  Technologies, Volume 2 (Short Papers)}, pages 107--112, New Orleans,
  Louisiana. Association for Computational Linguistics.

\bibitem[{Hao et~al.(2019)Hao, Dong, Wei, and Xu}]{Hao_Visualizing}
Yaru Hao, Li~Dong, Furu Wei, and Ke~Xu. 2019.
\newblock \href {https://doi.org/10.18653/v1/D19-1424} {Visualizing and
  understanding the effectiveness of {BERT}}.
\newblock In \emph{Proceedings of the 2019 Conference on Empirical Methods in
  Natural Language Processing and the 9th International Joint Conference on
  Natural Language Processing (EMNLP-IJCNLP)}, pages 4141--4150, Hong Kong,
  China. Association for Computational Linguistics.

\bibitem[{Hendrickx et~al.(2010)Hendrickx, Kim, Kozareva, Nakov,
  {\'O}~S{\'e}aghdha, Pad{\'o}, Pennacchiotti, Romano, and
  Szpakowicz}]{Hendrickx_Semeval}
Iris Hendrickx, Su~Nam Kim, Zornitsa Kozareva, Preslav Nakov, Diarmuid
  {\'O}~S{\'e}aghdha, Sebastian Pad{\'o}, Marco Pennacchiotti, Lorenza Romano,
  and Stan Szpakowicz. 2010.
\newblock \href {https://www.aclweb.org/anthology/S10-1006} {{S}em{E}val-2010
  task 8: Multi-way classification of semantic relations between pairs of
  nominals}.
\newblock In \emph{Proceedings of the 5th International Workshop on Semantic
  Evaluation}, pages 33--38, Uppsala, Sweden. Association for Computational
  Linguistics.

\bibitem[{Hewitt and Liang(2019)}]{Hewitt_Designing}
John Hewitt and Percy Liang. 2019.
\newblock \href {https://doi.org/10.18653/v1/D19-1275} {Designing and
  interpreting probes with control tasks}.
\newblock In \emph{Proceedings of the 2019 Conference on Empirical Methods in
  Natural Language Processing and the 9th International Joint Conference on
  Natural Language Processing (EMNLP-IJCNLP)}, pages 2733--2743, Hong Kong,
  China. Association for Computational Linguistics.

\bibitem[{Hewitt and Manning(2019)}]{Hewitt_A_Structural}
John Hewitt and Christopher~D. Manning. 2019.
\newblock \href {https://doi.org/10.18653/v1/N19-1419} {{A} structural probe
  for finding syntax in word representations}.
\newblock In \emph{Proceedings of the 2019 Conference of the North {A}merican
  Chapter of the Association for Computational Linguistics: Human Language
  Technologies, Volume 1 (Long and Short Papers)}, pages 4129--4138,
  Minneapolis, Minnesota. Association for Computational Linguistics.

\bibitem[{Hoover et~al.(2019)Hoover, Strobelt, and Gehrmann}]{Hoover_Exbert}
Benjamin Hoover, Hendrik Strobelt, and Sebastian Gehrmann. 2019.
\newblock exbert: A visual analysis tool to explore learned representations in
  transformers models.
\newblock \emph{arXiv preprint arXiv:1910.05276}.

\bibitem[{Jain and Wallace(2019)}]{Jain_Attention}
Sarthak Jain and Byron~C. Wallace. 2019.
\newblock \href {https://doi.org/10.18653/v1/N19-1357} {{A}ttention is not
  {E}xplanation}.
\newblock In \emph{Proceedings of the 2019 Conference of the North {A}merican
  Chapter of the Association for Computational Linguistics: Human Language
  Technologies, Volume 1 (Long and Short Papers)}, pages 3543--3556,
  Minneapolis, Minnesota. Association for Computational Linguistics.

\bibitem[{Jawahar et~al.(2019)Jawahar, Sagot, and Seddah}]{Jawahar_What}
Ganesh Jawahar, Beno{\^\i}t Sagot, and Djam{\'e} Seddah. 2019.
\newblock \href {https://doi.org/10.18653/v1/P19-1356} {What does {BERT} learn
  about the structure of language?}
\newblock In \emph{Proceedings of the 57th Annual Meeting of the Association
  for Computational Linguistics}, pages 3651--3657, Florence, Italy.
  Association for Computational Linguistics.

\bibitem[{Jia and Liang(2017)}]{Jia_Adversarial}
Robin Jia and Percy Liang. 2017.
\newblock \href {https://doi.org/10.18653/v1/D17-1215} {Adversarial examples
  for evaluating reading comprehension systems}.
\newblock In \emph{Proceedings of the 2017 Conference on Empirical Methods in
  Natural Language Processing}, pages 2021--2031, Copenhagen, Denmark.
  Association for Computational Linguistics.

\bibitem[{Kim et~al.(2018)Kim, Malon, and Kadav}]{Kim_Teaching}
Juho Kim, Christopher Malon, and Asim Kadav. 2018.
\newblock \href {https://doi.org/10.18653/v1/W18-5512} {Teaching syntax by
  adversarial distraction}.
\newblock In \emph{Proceedings of the First Workshop on Fact Extraction and
  {VER}ification ({FEVER})}, pages 79--84, Brussels, Belgium. Association for
  Computational Linguistics.

\bibitem[{Kingma and Ba(2014)}]{Kingma_Adam}
Diederik Kingma and Jimmy Ba. 2014.
\newblock Adam: A method for stochastic optimization.
\newblock \emph{International Conference on Learning Representations}.

\bibitem[{Kriegeskorte et~al.(2008)Kriegeskorte, Mur, and
  Bandettini}]{Kriegeskorte_Representational}
N.~Kriegeskorte, M.~Mur, and P.~Bandettini. 2008.
\newblock {{R}epresentational similarity analysis - connecting the branches of
  systems neuroscience}.
\newblock \emph{Front Syst Neurosci}, 2:4.

\bibitem[{Laakso and Cottrell(2000)}]{Laakso_Content}
Aarre Laakso and Garrison Cottrell. 2000.
\newblock Content and cluster analysis: assessing representational similarity
  in neural systems.
\newblock \emph{Philosophical psychology}, 13(1):47--76.

\bibitem[{Lan et~al.(2019)Lan, Chen, Goodman, Gimpel, Sharma, and
  Soricut}]{Anonymous_ALBERT}
Zhenzhong Lan, Mingda Chen, Sebastian Goodman, Kevin Gimpel, Piyush Sharma, and
  Radu Soricut. 2019.
\newblock \href {http://arxiv.org/abs/1909.11942} {Albert: A lite bert for
  self-supervised learning of language representations}.

\bibitem[{Liu et~al.(2019)Liu, Gardner, Belinkov, Peters, and
  Smith}]{Liu_Linguistic}
Nelson~F. Liu, Matt Gardner, Yonatan Belinkov, Matthew~E. Peters, and Noah~A.
  Smith. 2019.
\newblock \href {https://doi.org/10.18653/v1/N19-1112} {Linguistic knowledge
  and transferability of contextual representations}.
\newblock In \emph{Proceedings of the 2019 Conference of the North {A}merican
  Chapter of the Association for Computational Linguistics: Human Language
  Technologies, Volume 1 (Long and Short Papers)}, pages 1073--1094,
  Minneapolis, Minnesota. Association for Computational Linguistics.

\bibitem[{Marvin and Linzen(2018)}]{Marvin_Targeted}
Rebecca Marvin and Tal Linzen. 2018.
\newblock \href {https://doi.org/10.18653/v1/D18-1151} {Targeted syntactic
  evaluation of language models}.
\newblock In \emph{Proceedings of the 2018 Conference on Empirical Methods in
  Natural Language Processing}, pages 1192--1202, Brussels, Belgium.
  Association for Computational Linguistics.

\bibitem[{McCoy et~al.(2019{\natexlab{a}})McCoy, Min, and Linzen}]{Mccoy_Berts}
R~Thomas McCoy, Junghyun Min, and Tal Linzen. 2019{\natexlab{a}}.
\newblock Berts of a feather do not generalize together: Large variability in
  generalization across models with similar test set performance.
\newblock \emph{arXiv preprint arXiv:1911.02969}.

\bibitem[{McCoy et~al.(2019{\natexlab{b}})McCoy, Pavlick, and
  Linzen}]{Mccoy_Right}
Tom McCoy, Ellie Pavlick, and Tal Linzen. 2019{\natexlab{b}}.
\newblock \href {https://doi.org/10.18653/v1/P19-1334} {Right for the wrong
  reasons: Diagnosing syntactic heuristics in natural language inference}.
\newblock In \emph{Proceedings of the 57th Annual Meeting of the Association
  for Computational Linguistics}, pages 3428--3448, Florence, Italy.
  Association for Computational Linguistics.

\bibitem[{Morcos et~al.(2018)Morcos, Raghu, and Bengio}]{Morcos_Insights}
Ari Morcos, Maithra Raghu, and Samy Bengio. 2018.
\newblock \href
  {http://papers.nips.cc/paper/7815-insights-on-representational-similarity-in-neural-networks-with-canonical-correlation.pdf}
  {Insights on representational similarity in neural networks with canonical
  correlation}.
\newblock In S.~Bengio, H.~Wallach, H.~Larochelle, K.~Grauman, N.~Cesa-Bianchi,
  and R.~Garnett, editors, \emph{Advances in Neural Information Processing
  Systems 31}, pages 5727--5736. Curran Associates, Inc.

\bibitem[{Peters et~al.(2018{\natexlab{a}})Peters, Neumann, Iyyer, Gardner,
  Clark, Lee, and Zettlemoyer}]{Peters_Deep}
Matthew Peters, Mark Neumann, Mohit Iyyer, Matt Gardner, Christopher Clark,
  Kenton Lee, and Luke Zettlemoyer. 2018{\natexlab{a}}.
\newblock \href {https://doi.org/10.18653/v1/N18-1202} {Deep contextualized
  word representations}.
\newblock In \emph{Proceedings of the 2018 Conference of the North {A}merican
  Chapter of the Association for Computational Linguistics: Human Language
  Technologies, Volume 1 (Long Papers)}, pages 2227--2237, New Orleans,
  Louisiana. Association for Computational Linguistics.

\bibitem[{Peters et~al.(2018{\natexlab{b}})Peters, Neumann, Zettlemoyer, and
  Yih}]{Peters_Dissecting}
Matthew Peters, Mark Neumann, Luke Zettlemoyer, and Wen-tau Yih.
  2018{\natexlab{b}}.
\newblock \href {https://doi.org/10.18653/v1/D18-1179} {Dissecting contextual
  word embeddings: Architecture and representation}.
\newblock In \emph{Proceedings of the 2018 Conference on Empirical Methods in
  Natural Language Processing}, pages 1499--1509, Brussels, Belgium.
  Association for Computational Linguistics.

\bibitem[{Peters et~al.(2019)Peters, Ruder, and Smith}]{Peters_Tune}
Matthew~E. Peters, Sebastian Ruder, and Noah~A. Smith. 2019.
\newblock \href {https://doi.org/10.18653/v1/W19-4302} {To tune or not to tune?
  adapting pretrained representations to diverse tasks}.
\newblock In \emph{Proceedings of the 4th Workshop on Representation Learning
  for NLP (RepL4NLP-2019)}, pages 7--14, Florence, Italy. Association for
  Computational Linguistics.

\bibitem[{Poliak et~al.(2018)Poliak, Naradowsky, Haldar, Rudinger, and
  Van~Durme}]{Poliak_Hypothesis}
Adam Poliak, Jason Naradowsky, Aparajita Haldar, Rachel Rudinger, and Benjamin
  Van~Durme. 2018.
\newblock \href {https://doi.org/10.18653/v1/S18-2023} {Hypothesis only
  baselines in natural language inference}.
\newblock In \emph{Proceedings of the Seventh Joint Conference on Lexical and
  Computational Semantics}, pages 180--191, New Orleans, Louisiana. Association
  for Computational Linguistics.

\bibitem[{Raffel et~al.(2019)Raffel, Shazeer, Roberts, Lee, Narang, Matena,
  Zhou, Li, and Liu}]{Raffel_Exploring}
Colin Raffel, Noam Shazeer, Adam Roberts, Katherine Lee, Sharan Narang, Michael
  Matena, Yanqi Zhou, Wei Li, and Peter~J Liu. 2019.
\newblock Exploring the limits of transfer learning with a unified text-to-text
  transformer.
\newblock \emph{arXiv preprint arXiv:1910.10683}.

\bibitem[{Rajpurkar et~al.(2016)Rajpurkar, Zhang, Lopyrev, and
  Liang}]{Rajpurkar_SQuAD}
Pranav Rajpurkar, Jian Zhang, Konstantin Lopyrev, and Percy Liang. 2016.
\newblock \href {https://doi.org/10.18653/v1/D16-1264} {{SQ}u{AD}: 100,000+
  questions for machine comprehension of text}.
\newblock In \emph{Proceedings of the 2016 Conference on Empirical Methods in
  Natural Language Processing}, pages 2383--2392, Austin, Texas. Association
  for Computational Linguistics.

\bibitem[{Rogers et~al.(2020)Rogers, Kovaleva, and Rumshisky}]{Rogers_Primer}
Anna Rogers, Olga Kovaleva, and Anna Rumshisky. 2020.
\newblock A primer in bertology: What we know about how bert works.
\newblock \emph{arXiv preprint arXiv:2002.12327}.

\bibitem[{Saphra and Lopez(2019)}]{Saphra_Understanding}
Naomi Saphra and Adam Lopez. 2019.
\newblock \href {https://doi.org/10.18653/v1/N19-1329} {Understanding learning
  dynamics of language models with {SVCCA}}.
\newblock In \emph{Proceedings of the 2019 Conference of the North {A}merican
  Chapter of the Association for Computational Linguistics: Human Language
  Technologies, Volume 1 (Long and Short Papers)}, pages 3257--3267,
  Minneapolis, Minnesota. Association for Computational Linguistics.

\bibitem[{Serrano and Smith(2019)}]{Serrano_Is}
Sofia Serrano and Noah~A. Smith. 2019.
\newblock \href {https://doi.org/10.18653/v1/P19-1282} {Is attention
  interpretable?}
\newblock In \emph{Proceedings of the 57th Annual Meeting of the Association
  for Computational Linguistics}, pages 2931--2951, Florence, Italy.
  Association for Computational Linguistics.

\bibitem[{Silveira et~al.(2014)Silveira, Dozat, de~Marneffe, Bowman, Connor,
  Bauer, and Manning}]{Silveira_A}
Natalia Silveira, Timothy Dozat, Marie-Catherine de~Marneffe, Samuel Bowman,
  Miriam Connor, John Bauer, and Christopher~D. Manning. 2014.
\newblock A gold standard dependency corpus for {E}nglish.
\newblock In \emph{Proceedings of the Ninth International Conference on
  Language Resources and Evaluation (LREC-2014)}.

\bibitem[{Teichert et~al.(2017)Teichert, Poliak, Van~Durme, and
  Gormley}]{Teichert_Semantic}
Adam Teichert, Adam Poliak, Benjamin Van~Durme, and Matthew~R Gormley. 2017.
\newblock Semantic proto-role labeling.
\newblock In \emph{Thirty-First AAAI Conference on Artificial Intelligence
  (AAAI-17)}.

\bibitem[{Tenney et~al.(2019{\natexlab{a}})Tenney, Das, and
  Pavlick}]{Tenney_BERT}
Ian Tenney, Dipanjan Das, and Ellie Pavlick. 2019{\natexlab{a}}.
\newblock \href {https://doi.org/10.18653/v1/P19-1452} {{BERT} rediscovers the
  classical {NLP} pipeline}.
\newblock In \emph{Proceedings of the 57th Annual Meeting of the Association
  for Computational Linguistics}, pages 4593--4601, Florence, Italy.
  Association for Computational Linguistics.

\bibitem[{Tenney et~al.(2019{\natexlab{b}})Tenney, Xia, Chen, Wang, Poliak,
  McCoy, Kim, Durme, Bowman, Das, and Pavlick}]{Tenney_What}
Ian Tenney, Patrick Xia, Berlin Chen, Alex Wang, Adam Poliak, R~Thomas McCoy,
  Najoung Kim, Benjamin~Van Durme, Sam Bowman, Dipanjan Das, and Ellie Pavlick.
  2019{\natexlab{b}}.
\newblock \href {https://openreview.net/forum?id=SJzSgnRcKX} {What do you learn
  from context? probing for sentence structure in contextualized word
  representations}.
\newblock In \emph{International Conference on Learning Representations}.

\bibitem[{Vaswani et~al.(2017)Vaswani, Shazeer, Parmar, Uszkoreit, Jones,
  Gomez, Kaiser, and Polosukhin}]{Vaswani_Attention}
Ashish Vaswani, Noam Shazeer, Niki Parmar, Jakob Uszkoreit, Llion Jones,
  Aidan~N Gomez, {\L}ukasz Kaiser, and Illia Polosukhin. 2017.
\newblock Attention is all you need.
\newblock In \emph{Advances in neural information processing systems}, pages
  5998--6008.

\bibitem[{Vig and Belinkov(2019)}]{Vig_Analyzing}
Jesse Vig and Yonatan Belinkov. 2019.
\newblock \href {https://doi.org/10.18653/v1/W19-4808} {Analyzing the structure
  of attention in a transformer language model}.
\newblock In \emph{Proceedings of the 2019 ACL Workshop BlackboxNLP: Analyzing
  and Interpreting Neural Networks for NLP}, pages 63--76, Florence, Italy.
  Association for Computational Linguistics.

\bibitem[{Voita et~al.(2019{\natexlab{a}})Voita, Sennrich, and
  Titov}]{Voita_The_Bottom}
Elena Voita, Rico Sennrich, and Ivan Titov. 2019{\natexlab{a}}.
\newblock \href {https://doi.org/10.18653/v1/D19-1448} {The bottom-up evolution
  of representations in the transformer: A study with machine translation and
  language modeling objectives}.
\newblock In \emph{Proceedings of the 2019 Conference on Empirical Methods in
  Natural Language Processing and the 9th International Joint Conference on
  Natural Language Processing (EMNLP-IJCNLP)}, pages 4395--4405, Hong Kong,
  China. Association for Computational Linguistics.

\bibitem[{Voita et~al.(2019{\natexlab{b}})Voita, Talbot, Moiseev, Sennrich, and
  Titov}]{Voita_Analyzing}
Elena Voita, David Talbot, Fedor Moiseev, Rico Sennrich, and Ivan Titov.
  2019{\natexlab{b}}.
\newblock \href {https://doi.org/10.18653/v1/P19-1580} {Analyzing multi-head
  self-attention: Specialized heads do the heavy lifting, the rest can be
  pruned}.
\newblock In \emph{Proceedings of the 57th Annual Meeting of the Association
  for Computational Linguistics}, pages 5797--5808, Florence, Italy.
  Association for Computational Linguistics.

\bibitem[{Wallace et~al.(2019)Wallace, Wang, Li, Singh, and
  Gardner}]{Wallace_Do}
Eric Wallace, Yizhong Wang, Sujian Li, Sameer Singh, and Matt Gardner. 2019.
\newblock \href {https://doi.org/10.18653/v1/D19-1534} {Do {NLP} models know
  numbers? probing numeracy in embeddings}.
\newblock In \emph{Proceedings of the 2019 Conference on Empirical Methods in
  Natural Language Processing and the 9th International Joint Conference on
  Natural Language Processing (EMNLP-IJCNLP)}, pages 5306--5314, Hong Kong,
  China. Association for Computational Linguistics.

\bibitem[{Wang et~al.(2019)Wang, Singh, Michael, Hill, Levy, and
  Bowman}]{Wang_GLUE}
Alex Wang, Amanpreet Singh, Julian Michael, Felix Hill, Omer Levy, and
  Samuel~R. Bowman. 2019.
\newblock \href {https://openreview.net/forum?id=rJ4km2R5t7} {{GLUE}: A
  multi-task benchmark and analysis platform for natural language
  understanding}.
\newblock In \emph{International Conference on Learning Representations}.

\bibitem[{Weischedel et~al.(2013)Weischedel, Palmer, Marcus, Hovy, Pradhan,
  Ramshaw, Xue, Taylor, Kaufman, Franchini et~al.}]{Weischedel_Ontonotes}
Ralph Weischedel, Martha Palmer, Mitchell Marcus, Eduard Hovy, Sameer Pradhan,
  Lance Ramshaw, Nianwen Xue, Ann Taylor, Jeff Kaufman, Michelle Franchini,
  et~al. 2013.
\newblock Ontonotes release 5.0 ldc2013t19.
\newblock \emph{Linguistic Data Consortium, Philadelphia, PA}, 23.

\bibitem[{Williams et~al.(2018)Williams, Nangia, and Bowman}]{Williams_MNLI}
Adina Williams, Nikita Nangia, and Samuel Bowman. 2018.
\newblock \href {http://aclweb.org/anthology/N18-1101} {A broad-coverage
  challenge corpus for sentence understanding through inference}.
\newblock In \emph{Proceedings of the 2018 Conference of the North American
  Chapter of the Association for Computational Linguistics: Human Language
  Technologies, Volume 1 (Long Papers)}, pages 1112--1122. Association for
  Computational Linguistics.

\bibitem[{Wu et~al.(2016)Wu, Schuster, Chen, Le, Norouzi, Macherey, Krikun,
  Cao, Gao, Macherey, Klingner, Shah, Johnson, Liu, Łukasz Kaiser, Gouws,
  Kato, Kudo, Kazawa, Stevens, Kurian, Patil, Wang, Young, Smith, Riesa,
  Rudnick, Vinyals, Corrado, Hughes, and Dean}]{Wu_Googles}
Yonghui Wu, Mike Schuster, Zhifeng Chen, Quoc~V. Le, Mohammad Norouzi, Wolfgang
  Macherey, Maxim Krikun, Yuan Cao, Qin Gao, Klaus Macherey, Jeff Klingner,
  Apurva Shah, Melvin Johnson, Xiaobing Liu, Łukasz Kaiser, Stephan Gouws,
  Yoshikiyo Kato, Taku Kudo, Hideto Kazawa, Keith Stevens, George Kurian,
  Nishant Patil, Wei Wang, Cliff Young, Jason Smith, Jason Riesa, Alex Rudnick,
  Oriol Vinyals, Greg Corrado, Macduff Hughes, and Jeffrey Dean. 2016.
\newblock \href {http://arxiv.org/abs/1609.08144} {Google's neural machine
  translation system: Bridging the gap between human and machine translation}.

\bibitem[{Zeman et~al.(2017)Zeman, Popel, Straka, Hajic, Nivre, Ginter,
  Luotolahti, Pyysalo, Petrov, Potthast, Tyers, Badmaeva, Gokirmak, Nedoluzhko,
  Cinkova, Hajic~jr., Hlavacova, Kettnerov\'{a}, Uresova, Kanerva, Ojala,
  Missil\"{a}, Manning, Schuster, Reddy, Taji, Habash, Leung, de~Marneffe,
  Sanguinetti, Simi, Kanayama, dePaiva, Droganova, Mart\'{i}nez~Alonso,
  \c{C}\"{o}ltekin, Sulubacak, Uszkoreit, Macketanz, Burchardt, Harris,
  Marheinecke, Rehm, Kayadelen, Attia, Elkahky, Yu, Pitler, Lertpradit, Mandl,
  Kirchner, Alcalde, Strnadov\'{a}, Banerjee, Manurung, Stella, Shimada, Kwak,
  Mendonca, Lando, Nitisaroj, and Li}]{Zeman_CoNLL}
Daniel Zeman, Martin Popel, Milan Straka, Jan Hajic, Joakim Nivre, Filip
  Ginter, Juhani Luotolahti, Sampo Pyysalo, Slav Petrov, Martin Potthast,
  Francis Tyers, Elena Badmaeva, Memduh Gokirmak, Anna Nedoluzhko, Silvie
  Cinkova, Jan Hajic~jr., Jaroslava Hlavacova, V\'{a}clava Kettnerov\'{a},
  Zdenka Uresova, Jenna Kanerva, Stina Ojala, Anna Missil\"{a}, Christopher~D.
  Manning, Sebastian Schuster, Siva Reddy, Dima Taji, Nizar Habash, Herman
  Leung, Marie-Catherine de~Marneffe, Manuela Sanguinetti, Maria Simi, Hiroshi
  Kanayama, Valeria dePaiva, Kira Droganova, H\'{e}ctor Mart\'{i}nez~Alonso,
  \c{C}a\u{g}rı \c{C}\"{o}ltekin, Umut Sulubacak, Hans Uszkoreit, Vivien
  Macketanz, Aljoscha Burchardt, Kim Harris, Katrin Marheinecke, Georg Rehm,
  Tolga Kayadelen, Mohammed Attia, Ali Elkahky, Zhuoran Yu, Emily Pitler, Saran
  Lertpradit, Michael Mandl, Jesse Kirchner, Hector~Fernandez Alcalde, Jana
  Strnadov\'{a}, Esha Banerjee, Ruli Manurung, Antonio Stella, Atsuko Shimada,
  Sookyoung Kwak, Gustavo Mendonca, Tatiana Lando, Rattima Nitisaroj, and Josie
  Li. 2017.
\newblock \href {http://www.aclweb.org/anthology/K/K17/K17-3001.pdf} {Conll
  2017 shared task: Multilingual parsing from raw text to universal
  dependencies}.
\newblock In \emph{Proceedings of the CoNLL 2017 Shared Task: Multilingual
  Parsing from Raw Text to Universal Dependencies}, pages 1--19, Vancouver,
  Canada. Association for Computational Linguistics.

\bibitem[{Zhang and Bowman(2018)}]{Zhang_Language}
Kelly~W Zhang and Samuel~R Bowman. 2018.
\newblock Language modeling teaches you more syntax than translation does:
  Lessons learned through auxiliary task analysis.
\newblock \emph{arXiv preprint arXiv:1809.10040}.

\bibitem[{{Zhu} et~al.(2015){Zhu}, {Kiros}, {Zemel}, {Salakhutdinov},
  {Urtasun}, {Torralba}, and {Fidler}}]{Zhu_Aligning}
Y.~{Zhu}, R.~{Kiros}, R.~{Zemel}, R.~{Salakhutdinov}, R.~{Urtasun},
  A.~{Torralba}, and S.~{Fidler}. 2015.
\newblock \href {https://doi.org/10.1109/ICCV.2015.11} {Aligning books and
  movies: Towards story-like visual explanations by watching movies and reading
  books}.
\newblock In \emph{2015 IEEE International Conference on Computer Vision
  (ICCV)}, pages 19--27.

\end{thebibliography}
\bibliographystyle{acl_natbib}

\end{document}